Preprint Notice:

# Barking up the Right Tree:
# Improving Cross-Corpus Speech Emotion Recognition with Adversarial Discriminative Domain Generalization (ADDoG)

John Gideon, *Member, IEEE,* Melvin G McInnis, and Emily Mower Provost, *Senior Member, IEEE,*

**Abstract**—Automatic speech emotion recognition provides computers with critical context to enable user understanding. While methods trained and tested within the same dataset have been shown successful, they often fail when applied to unseen datasets. To address this, recent work has focused on adversarial methods to find more generalized representations of emotional speech. However, many of these methods have issues converging, and only involve datasets collected in laboratory conditions. In this paper, we introduce Adversarial Discriminative Domain Generalization (ADDoG), which follows an easier to train "meet in the middle" approach. The model iteratively moves representations learned for each dataset closer to one another, improving cross-dataset generalization. We also introduce Multiclass ADDoG, or MADDoG, which is able to extend the proposed method to more than two datasets, simultaneously. Our results show consistent convergence for the introduced methods, with significantly improved results when not using labels from the target dataset. We also show how, in most cases, ADDoG and MADDoG can be used to improve upon baseline state-of-the-art methods when target dataset labels are added and in-the-wild data are considered. Even though our experiments focus on cross-corpus speech emotion, these methods could be used to remove unwanted factors of variation in other settings.

✦

## 1 INTRODUCTION

AUTOMATIC speech emotion recognition has been explored by researchers as a way to bridge the communication gap between humans and computers. This information is critical to sustain long-term interaction between humans and machines. For example, emotion can impact a person's current attention level, their ability to perceive and remember information, and their capacity to make decisions [1]. However, compared with many other machine learning tasks, including automatic speech recognition (ASR), speech emotion datasets are much smaller and less varied [2]. This is due to the large time and effort often needed to collect and annotate emotional speech. As a result, even when an emotion model is successfully trained on one dataset, it often fails when applied to another [3]. This has motivated researchers to explore cross-corpus training methods to be able to utilize multiple datasets at once and to create systems more robust to unseen data. In this paper, we introduce and explore new methods for generalizing representations of speech for emotion so that recognition performance is improved across datasets.

Emotion is only one of several factors that impacts the acoustics of speech. Some factors that change across datasets and can impact affect recognition include the environmental noise [4], the spoken language [3], the recording device quality [5], and the elicitation strategy (acted versus natural) [6]. Additionally, a mismatch in subject demographics between datasets can result in misclassification, due to the small numbers of participants common in speech emotion recognition datasets

• J. Gideon, M.G. McInnis, and E. Mower Provost are with the University of Michigan, Ann Arbor, MI, USA.
  E-mail: {gideonjn, mmcinnis, emilykmp}@umich.edu

[2]. Early work in cross-corpus speech emotion recognition attempted to address these differences with feature normalization [3], [7], sample selection [8], and decision fusion [9]. Most modern techniques of cross-corpus speech emotion recognition use deep learning to build representations over low-level acoustic features. Many of these techniques incorporate tasks in addition to emotion to be able to learn more robust representations [6], [10].

More recently, speech research has followed the popularization of adversarial methods, including Generative Adversarial Networks (GANs) [11], [12], [13], [14], Wassersteain GANS (WGANS) [15], and CycleGANs [16], [17], [18]. However, many of these generative speech transfer models introduce noise and have a long way to go, as explored by Kaneko et al. [18]. To get around this issue, some cross-corpus research has instead explored discriminative adversarial methods, including Adversarial Discriminative Domain Adaptation (ADDA) [19] and Domain Adversarial Neural Networks (DANNs) [20], [21]. While ADDA has seen effective application in image recognition [19], it has not yet been successfully applied to speech emotion. This is likely because the target representation is learned independent of the output classifier and there is no guarantee that a lower varying characteristic, like emotion, would be preserved. DANNs work by using the GAN discriminator with the aim to "unlearn" domain from a target representation [20]. While this has been effectively applied to emotion, the authors note the difficulty in getting the method to converge in certain cases [21]. All of these methods are explained in greater detail in Section 2.

In this paper, we investigate three models for speech emotion recognition across datasets. Our baseline model is a Convolutional Neural Network (CNN), which is commonly employed in automatic speech emotion recognition [22], [23], [24]. CNNs are able to learn temporal filters across features and distill an entire utterance down into a static representation for more conventional fully connected layers to model. However, this model does not explicitly capture the effects of dataset or incorporate unlabelled data. We introduce Adversarial Discriminative Domain Generalization (ADDoG) - a method for finding a more generalized intermediate representation for speech emotion across datasets. The network implementation is similar to CNN, except that an additional critic network is appended to the utterance-level representation layer. We adversarially train this network to iteratively move the different dataset representations closer to one another. We demonstrate that this "meet in the middle" approach always converges and improves upon previous, less stable, methods [20], [21]. Finally, we implement and explore Multiclass ADDoG, or MADDoG, which is able to incorporate many datasets at a time and build an even more robust and generalized representation.

We propose four sets of experiments to determine the effectiveness of our models for cross-corpus testing under different conditions. Our first experiment examines the case of training on one dataset and testing on another, allowing ADDoG to also incorporate the unlabelled test features for training. This mirrors the transductive learning approach, seen in prior speech emotion work [25], [26]. We constrain our first experiment to only consider datasets recorded in a laboratory environment. Experiment 2 expands on this by introducing increasing amounts of labelled data available from the target dataset. Experiment 3 explores the impact of training on a laboratory dataset and testing on an in-the-wild dataset. We also look into incorporating three total datasets into training simultaneously, and present MADDoG as especially suited to this problem. Finally, Experiment 4 does the reverse of Experiment 3 and investigates training on in-the-wild data and testing on more traditionally recorded laboratory speech.

Our results indicate that ADDoG consistently converges and is able to construct a more generalized representation for cross-corpus testing. This affirms the iterative "meet in the middle" approach to domain generalization. We find significant improvements in performance versus the baseline systems in all experiments with no added labelled target data. In addition to attaining higher performance, the ADDoG results have lower variance across repeated experiments, indicating better stability, when compared with CNN. Additional experiments show that ADDoG performs the best in the majority of cases when labelled target data is available, especially when the set is fairly small. However, the margin of improvement decreases with more added target data, implying that there is a trade-off between building a generalized model and specializing to the target domain. This trend holds true even with in-the-wild target data, demonstrating the robustness of the ADDoG technique. We find the improvement in performance to be at least as good as the benefit of doubling the amount of labelled target data. Finally, we show that MADDoG is able to improve further upon ADDoG when multiple source datasets are available by explicitly modelling all dataset differences.

The novelty of this paper includes: (1) the ADDoG model, which allows for better generalized representation convergence by "meeting in the middle"; (2) the MADDoG method, which extends ADDoG to allow for many dataset differences to be explicitly modelled; (3) an analysis of cross-corpus experiments where both laboratory and in-the-wild data are trained and then tested on the other.

Fig. 1. The main domain adaptation and domain generalization methods referenced in this paper, divided by generative and discriminative methods. Prior work is listed with related citations and abbreviations defined in Section 2. Methods introduced in this paper are bolded and are explained in Section 4.

## 2 RELATED WORKS

### 2.1 Cross-Corpus Speech Emotion Recognition

Speech emotion models trained and tested on a single dataset often fail when new datasets are introduced. This can be due to differences in recording conditions, microphone quality, elicitation strategy (acted versus natural), and the distribution of labels [2]. Additionally, the demographics of subjects may widely vary between datasets [2]. Figure 1 gives a categorization of the main methods referenced in this paper.

One of the earliest works in cross-corpus emotion by Schuller et al. examined how acoustic and annotation differences can result in decreased performance [3]. They explored different techniques of feature normalization and found speaker-based z-normalization to work best. Additionally, they demonstrated how differences in selected sub-groups of emotions can cause large discrepancies in performance. This indicated the importance of carefully selecting annotations across all datasets in a multi-corpus experiment. Zhang et al. addressed the problem of dataset label mismatch by creating a knowledge-based mapping between classes [7]. They further explored feature normalization for utterance-level features and found that within-corpus normalization with unlabelled data boosted performance. Additional work by Schuller et al. explored how selecting only the most prototypical examples when training cross-dataset systems can improve activation classification [8]. This suggests that the most exemplar samples within datasets may also be those samples most consistently represented across datasets. Later work demonstrated how fusing the outputs of expert systems trained on individual datasets can outperform classifiers of the agglomerated data [9]. However, this performance difference depended heavily on the selected model.

While initial work focused on feature normalization, sample selection, and decision fusion, emotion researchers began exploring more complex methods of adapting features and models for cross-corpus testing. Hassan et al. explored how previous transfer learning methods, including Kernel Mean Matching (KMM), Unconstrained Least-Squares Importance Fitting (uLSIF), and the Kullback-Leibler Importance Estimation Procedure (KLIEP), could be used to compensate for dataset differences [32]. Song et al. investigated how dimensionality reduction algorithms could be used to form a similar feature space for different speech emotion datasets [33]. They found that Locality Preserving Projections (LPP), introduced in [34], resulted in the best classification performance. Abdelwahab et al. explored variations of support vector machines (SVMs) for supervised adaptation with small amounts of target domain data [35]. Using just 9% of the data, they were able to use Adaptive SVMs and Incremental SVMs to significantly improve cross-corpus performance compared with no data.

More recent speech emotion research has focused on learning deep representations over low-level features instead of the SVMs and utterance-level features found in prior work. In particular, training deep networks over multiple tasks simultaneously has been shown to improve performance when considering cross-corpus emotion. Parthasarathy et al. explored jointly predicting activation, valence, and dominance with a deep neural network (DNN), considering one as the primary task and the others as auxiliary [10]. This method significantly increased cross-corpus performance compared to a single task system, especially for models with large layer sizes. Kim et al. investigated whether adding additional non-emotion tasks, such as gender and the naturalness of the expression, would improve cross-corpus performance [6]. They achieved better or comparable performance when compared with systems not incorporating the additional tasks.

Another cross-domain framework that has been effectively applied within the speech emotion community is transductive learning. Traditional inductive learning first learns a model and then makes predictions on unseen data. Transductive learning instead aims to make predictions on a set of test data known in advance [36]. Because of this, it is possible to incorporate the unlabelled test data into the training procedure. Zong et al. explored an extension of linear discriminate analysis (LDA) for improving cross-corpus emotion recognition from speech [25]. Their method, called sparse transductive transfer LDA, or STTLDA, achieved significant improvement over SVM. Further work by Song et al. demonstrated another extension of LDA, Transfer Supervised Linear Subspace Learning (TSLSL), which again provided improvement for speech emotion within the transductive framework [26].

## 2.2 Adversarial Methods

With the introduction of Generative Adversarial Networks (GANs) [11], there has been a large increase in the amount of adversarial methods for cross-corpus modelling. GANs work by iteratively training a generator and a discriminator. The generator aims to create data that matches a certain distribution of real examples from just a random seed. The discriminator is trained to be able to tell apart these generated and real examples. A well-trained discriminator can be used to improve the authenticity of generated data by training a generator to fool the discriminator. To further expand on GANs, Radford et al. introduced Deep Convolutional GANs (DCGANs) [12]. They were able to improve convergence by using convolutions instead of fully connected layers. Automatic speech emotion recognition has begun to take advantage of these methods. Sahu et al. [13] explored how GANs could be used to augment utterance-level features, while Chang et al. [14] used DCGANs to improve performance on spectrograms.

The training of GANs can be unstable and relies on carefully tuned learning rates and numbers of generator versus discriminator iterations for convergence [12]. To address this issue, Arjovsky et al. introduced the Wasserstein GAN (WGAN) and was able to improve the convergence of GANs with a few minor changes [15]. The discriminator is replaced with a critic by removing the sigmoid activation on the output. Instead of trying to determine if examples are real or fake, like a discriminator, it instead learns to approximate the Wasserstein, or earth-mover's, distance between the real and fake distributions. This allows the system to have less sensitivity to over-training a discriminator, which often results in a saturated sigmoid function. It accomplishes this by clipping the critic weights to small values every iteration. This enforces the Lipschitz constraint and ensures that the output of the network does not grow infinitely. Instead, the network can only increase the output by finding the most succinct method of differentiating the real and fake examples. This allows a critic to be trained for many iterations before training the generator, resulting in more reliable gradients.

One of the first GAN-based methods to show promise for cross-domain applications was the CycleGAN by Zhu et al. [16]. This allowed for style transfer of images by converting from the style of one domain to another, while preserving the overall structure. A CycleGAN consists of two DCGANs working in tandem to convert from domain one to domain two and vice versa. Because CycleGANs can transfer in both directions, they are able to be trained with an additional reconstruction term that makes sure the overall structure of a transferred image is maintained. Zhu et al. demonstrated that is was possible to augment a facial emotion dataset by using CycleGANs to transfer between different emotions [17]. This allowed for training with balanced classes by transferring all utterances to each class, regardless of the original emotion. They found improved classification performance using this balacing method. Kaneko et al. explored subject conversion for speech using CycleGANs with some success, but found that there was still a large gap in quality for the real versus transferred samples [18]. We were unable to find published work on CycleGANs for speech emotion recognition, possibly because of this lack of transfer quality.

To work around this transfer quality issue, some adversarial methods have instead forgone generative methods for discriminative ones. For example, Tzeng et al. explored the method Adversarial Discriminative Domain Adaptation (ADDA) for transferring to a target domain [19]. In the first stage of learning, the source domain data is encoded to an intermediate representation using a series of convolutions. The representation is then passed through a classifier and both networks are optimized based on the available labels. Next, a separate encoder is trained for the target dataset, using a discriminator to ensure that the source and target representations are similar. Finally, the target encoder is appended to the classifier that was trained on the source data and target predictions are output. This method produced significantly better and more balanced class performance on cross-corpus testing of numerical image datasets. However, this method makes the assumption that the representation trained by the second encoder will still preserve structure meaningful to the classifier.

### 2.3 Domain Generalization

Most of the prior cross-domain methods focus on transferring from one domain representation to another (domain adaptation). Domain generalization instead focuses on creating a middle-ground representation for all data [31]. Domain generalization methods can be divided into generative methods (usually autoencoder based) and discriminative methods.

One common method for finding a domain generalized representation is an autoencoder. Autoencoders work by converting the original features into a more compressed representation using smaller layer sizes, sparsity, or other regularization methods. Deng et al. examined the use of denoising autoencoders (DAAs) for cross-domain speech emotion recognition [4], [27]. DAAs introduce noise to the input features and encourage the network to compress and reconstruct the features without the noise. This allows for the intermediate representation to discover a more noise robust representation that can work well across domains. Further work has examined different variations of autoencoders for speech emotion recognition, including Variational Autoencoders (VAE) [28], [29], Adversarial Autoencoders (AAE) [29], [30], [31], and Adversarial Variational Bayes (AVB) [29].

Another method that does not rely on autoencoders is Domain Adversarial Neural Networks (DANNs), introduced by Ajakan et al. [20]. DANNs have three main components: (1) feature extractor; (2) label classifier; (3) domain classifier. The input data is passed through the feature extraction layers. The representation is then fed to both the emotion classifier and the domain classifier. However, unlike multitask learning, where the domain is just another task, a reversal gradient layer is applied to the input of the domain classifier. This results in the network backpropagating a gradient to correctly classify the label but to incorrectly classify the domain, generalizing the intermediate representation. Abdelwahab et al. successfully applied this method to cross-corpus speech emotion recognition and showed significant improvement versus a model trained on the source data alone [21].

### 2.4 Open Challenges

While many papers have explored cross-corpus speech emotion recognition, there are many challenges remaining for the field. Adversarial methods, led by CycleGANs [16], have shown promise for directly converting speech between datasets. Once all utterances are converted to one domain, the differences should no longer need to be considered during further modelling. However, there is much noise introduced in the output, making this currently impractical [18]. Other generative methods, such as autoencoders and their many variants, get around this by instead just using the intermediate representation for classification. Yet, it is unclear that compressing the representation preserves the emotion component of the signal,

### TABLE 1
### Summary of Datasets

| | IEMOCAP | MSP-Improv | PRIORI Emotion |
|---|---|---|---|
| Subjects | 10 | 12 | 12 |
|   Male | 5 | 6 | 5 |
|   Female | 5 | 6 | 7 |
| Environment | Laboratory | Laboratory | Cell Phone Calls |
| Sample Rate | 16 kHz | 44.1 kHz | 8 kHz |
| Valence Scale | 1 - 5 | 1 - 5 | 1 - 9 |
|   Mean | 2.79 | 3.02 | 4.86 |
|   Std. | 0.99 | 1.06 | 1.12 |
| Hours | 12.4 | 9.6 | 21.7 |
|   Without Ties | 8.6 | 8.9 | 16.4 |
| Utterances | 10039 | 8438 | 11402 |
|   Without Ties | 6816 | 7852 | 8685 |
|   Low Valence | 3181 | 2160 | 2809 |
|   Mid Valence | 1641 | 2961 | 4779 |
|   High Valence | 1994 | 2731 | 1097 |
| Utterance Len. | | | |
|   Mean (sec.) | 4.5 | 4.1 | 6.8 |
|   Std. (sec.) | 3.1 | 2.9 | 4.4 |

particularly given emotion's relatively slowly varying nature. In fact, prior work has shown that when the speech signal is compressed, the emotion content can be lost (e.g., Principal Component Analysis (PCA) [37], [38]).

Other discriminative methods have been introduced to avoid these issues, including ADDA [19] and DANN [21]. ADDA relies on training a feature transformation for the target dataset to match the mid-level representation for the source dataset. Yet, again, there is no guarantee that this transformation will preserve the emotion information present in the original example because the emotion classifier is trained separately. This is related to a well known issue with GANs, known as mode collapse, which results in the generator converging to just a few convincing examples, regardless of the input [15]. Additionally, emotion is less likely to be preserved when simply matching representation distributions, due to its relatively lower variability when compared with the entire speech signal (again, see the issues with emotion and PCA [37], [38]). The DANN method improves on this by relying on a shared feature representation [20]. However, the researchers noted that DANN had issues converging on certain sets of parameters when training on speech emotion [21]. This could be due to the fact that DANNs attempt to "unlearn" domain, producing an unclear gradient. Finally, along with most other previously referenced papers in speech emotion recognition, the demonstration was on laboratory recorded datasets (IEMOCAP and MSP-Improv) rather than in-the-wild corpora. Further work is needed to incorporate more datasets simultaneously to improve generalization, as well as an exploration of the challenges of working with in-the-wild data.

## 3 DATASETS

This section describes the three different datasets included in this work: IEMOCAP [39], MSP-Improv [40], and the PRIORI Emotion [41]. The next three sections describe summaries of each of the included datasets. Table 1 describes the main attributes of each dataset. The final sections describe the emotion labeling and the audio preprocessing and feature extraction.

## 3.1 IEMOCAP

The "Interactive Emotional Dyadic MOtion Capture Database" (IEMOCAP) was created to explore the relationship between emotion, gestures, and speech. Ten actors (five male and five female) were recorded over five sessions. Each session consisted of a male and a female performing given either a series of scripts or improvisational scenarios. During the session, motion capture markers were attached to just one of the actors at a time. Once all scripts and improvisations were performed, the other actor was given the motion capture markers and the whole process was repeated. The audio was recorded using two high quality shotgun microphones at a 48 kHz sampling rate and later downsampled to 16 kHz.

The data were segmented by speaker turn, resulting in 10,039 total utterances (5,255 scripted turns, 4,784 improvised turns). Segments were then annotated for emotion, including valence and activation on a 1 to 5 scale. Between two and four annotations were performed per utterance. Further information about the IEMOCAP dataset can be found in [39].

## 3.2 MSP-Improv

The MSP-Improv dataset aims to capture more naturalistic emotion from improvised scenarios, while also partially controlling for lexical content. The collection involved a total of twelve actors (six male and six female). Like IEMOCAP, the dataset is split into six sessions, each including interactions between one male actor and one female actor. Each actor wore a collar microphone to record speech at 48 kHz (later downsampled to 44.1 kHz).

MSP-Improv controls for lexical content by including specific "target sentences" with fixed lexical content that can be embedded into different scenarios (i.e., angry, happy, sad, neutral). In each pair, one of the actors was tasked with ensuring that the target sentence was spoken in each scenario. Once all target sentences and scenarios were recorded, the actors switched roles and the second actor assumed this responsibility. Using this method, the researchers were able to control for lexical content, while still allowing for more natural emotion expression.

The data was divided into 652 target sentences, 4,381 improvised turns (the remainder of the improvised scenario, excluding the target sentence), 2,785 natural interactions (interactions between the actors in between recordings of the scenarios), and 620 read sentences (emotional readings of the target sentences). This totaled 8,438 utterances over 8.9 hours. These utterances were then annotated for emotion using crowd-sourcing on Amazon Mechanical Turk. Valence and activation were rated on a scale from 1 to 5. There is a minimum of five annotators per utterance up to a maximum of 50 (median of 5). Please refer to [40] for additional information about the MSP-Improv dataset.

## 3.3 PRIORI Emotion Dataset

The PRIORI Emotion dataset is an affect-annotated subset of the larger PRIORI (Predicting Individual Outcomes for Rapid Intervention) bipolar mood dataset, which includes smartphone calls from 51 patients and 9 healthy controls over the course of six-months to a year [5], [41], [42], [43]. The PRIORI Emotion dataset was obtained by first automatically segmenting the PRIORI dataset using the COMBO-SAD algorithm [44], as described in our prior work [5]. We consider only the 12 subjects who provided consent to have their calls manually annotated. Segments were chosen amongst the 12 subjects to provide diversity across both subjects and mood states. The researchers manually examined each selected segment before annotation to remove those that were inappropriate (identifiable information, no speech, etc.). Finally, the segments were annotated by 11 annotators on a 9-point Likert scale. Each segment received between 2 and 6 ratings.

The dataset contains 13,611 segments over 25.2 hours. We downsampled this dataset, selecting segments for which all assigned annotators were able to provide a rating, resulting in a dataset with 11,402 utterances over 21.7 hours. See Khorram et al. [41] for further information about the PRIORI Emotion dataset and [5], [42], [43] for the PRIORI bipolar mood dataset.

## 3.4 Emotion Labeling

Each of the datasets are segmented and rated on a dimensional scale for valence and activation by multiple annotators. We use these dimensional ratings for emotion, instead of discrete classes, as we hypothesize they are more consistently interpretable across datasets [45]. Further, we focus only on valence in this study, as our preliminary experiments did not show a benefit to domain generalization for activation.

We follow the method similar to [14] and [46] to convert the dataset annotator ratings into a three bin vector for soft classification. The middle bin consists of valence ratings equal to 3 for IEMOCAP and MSP-Improv and 5 for PRIORI Emotion. The other two bins are valence ratings that fall below or above this midpoint. Each vector is formed by counting each of the ratings belonging to the bins. These counts are then divided by the total number of ratings so that the vector sums to one. For example, if three IEMOCAP annotators gave the ratings of 3, 4, and 5, the soft vector representation would be [0.0, 0.33, 0.66]. However, unlike [46], we do not include utterances with no clear majority bin to make the analysis more straightforward. This matches other speech emotion work that only used majority agreement [47]. Additionally, this corresponds with the finding by Schuller et al. that prototypical examples are more useful for cross-corpus speech emotion recognition [8].

The IEMOCAP dataset includes 3,181, 1,641, and 1,994 low, medium, and high valence utterances, respectively. The MSP-Improv dataset includes 2,160 low, 2,961 middle, and 2,731 high valence utterances. The PRIORI dataset includes 2,809 low, 4,779 middle, and 1,097 high valence utterances.

## 3.5 Preprocessing and Feature Extraction

We downsample the audio between datasets to match. For experiments involving just IEMOCAP and MSP-Improv the sample rate is 16 kHz. If PRIORI data are involved then all data are downsampled to 8 kHz. The audio is then normalized to 0dB FSD using the SoX command line tool [48].

We then extract 40 dimensional Mel Filter Banks (MFBs) using the Kaldi speech recognition toolkit [49]. The default options are used - a povey window with frame length of 25 ms, frame shift of 10 ms, preemphasis coefficient of 0.97, low cutoff of 20 Hz, and outputting log filterbanks. Because this produces features of different lengths for each utterance, batches have their MFBs padded by zeros to the length of the longest utterance.

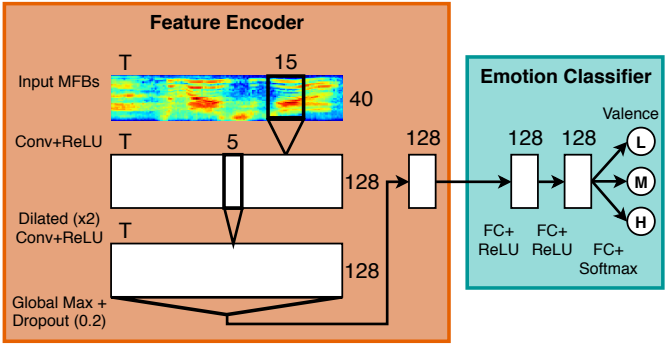

Fig. 2. Convolutional Neural Network (CNN). Consists of two main parts: (1) feature encoder; (2) emotion classifier. The feature encoder uses a set of convolutions and global pooling to create a 128-dimensional utterance level representation. The emotion classifier then uses fully connected layers and a softmax layer to output the three bin valence probability distribution.

## 4 CLASSIFICATION MODELS

In this section, we present the three different classification models used in this paper: a simple Convolutional Neural Network (CNN), Adversarial Discriminative Domain Generalization (ADDoG), and Multiclass ADDoG (MADDoG), which is an extension of ADDoG that allows for more than one source dataset. All models consider MFBs as the input feature set and valence binned into a three dimensional vector as the output task. Each experiment will consist of labelled data from a source dataset (SRC) and data from a target dataset (TAR), some of which is labelled and some is not. TAR contains the test data and is available at train time without labels (transductive learning). The baseline CNN method is able to take advantage of the labelled data from all datasets, but does not use unlabelled data. Both ADDoG and MADDoG take advantage of the unlabelled test data to generalize the intermediate feature representation across datasets. In all methods, we use the Adam optimizer [50] with default parameters ($\alpha = 0.0001, \beta_1 = 0.9, \beta_2 = 0.999$). All models described below are implemented in PyTorch version 0.4.0 [51].

### 4.1 CNN

Convolutional Neural Networks (CNN) have seen much success in speech emotion recognition [22], [23], [24]. Figure 2 shows our CNN implementation. It consists of two main components: (1) the feature encoder (convolutions + max pooling); (2) the emotion classifier (fully connected layers + softmax).

It is difficult to validate multiple sets of hyperparameters when conducting cross-dataset experiments, due to the lack of labelled data in the target domain. For this reason, we select hyperparameters based on those found to be commonly selected in prior work and keep them constant for all experiments. A channel size of 128 is used for all convolutional and fully connected layers, as commonly selected in prior work [24], [52]. ReLU is used as the activation function for all but the final layer, as it has been show successful in the field and is computationally efficient [24], [53]. We select a relatively large kernel size of 15 for the first convolutional layer, as previous work has shown large initial layers to be beneficial to emotion recognition using MFBs [23], [24]. We apply an additional convolutional layer of length 5 dilated by a factor of 2 to further extend the receptive field of the network. The global maximum is then taken over this convolution output, resulting in an encoded representation of 128 for the entire utterance. Previous work has shown that this is sufficient for recognizing emotion over short utterances [23], [24]. Dropout (p=0.2) is then applied to help prevent over-fitting. We next add three fully connected layers, with the final having three outputs for each of the valence bins, as in [46]. Finally, a softmax layer is applied, allowing for the output to be viewed as the probability distribution of valence. Biases are not used for any layers. Coupled with the ReLU activation and max pooling, this minimizes the effect of zero padding shorter utterances.

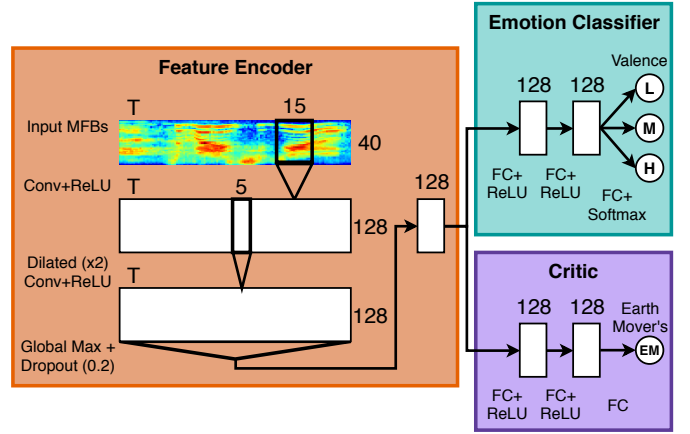

Fig. 3. Adversarial Discriminative Domain Generalization (ADDoG) Network. Consists of three main parts: (1) the feature encoder; (2) emotion classifier; (3) critic. The critic learns to estimate the earth mover's or Wasserstein distance between the SRC and TAR dataset encoded feature representations. The emotion classifier ensures that valence is also preserved in the generalized representation.

While older work in deep learning pretrained using autoencoders with unlabelled data, this has mostly subsided with the introduction of the ReLU activation, dropout, better initialization techniques, and larger datasets [54]. Because of this, our CNN model does not use the unlabelled data, and only the labelled data from both SRC and TAR is used during training. Each epoch is divided into a total number of batches equal to the amount of labelled data divided by the batch size. After the MFBs are propagated through the network, we calculate loss using a weighted cross entropy measure. The classes are weighted so that all valence bins are given equal likelihood, regardless of class imbalance.

### 4.2 ADDoG

We introduce Adversarial Discriminative Domain Generalization (ADDoG), which addresses the open challenges of producing a generalized dataset representation using unlabelled target data, while still being able to consistently converge. Similar to CNN, it builds an intermediate 128-dimensional encoding of the utterances after global max pooling and dropout. However, in the case of ADDoG, there is a critic component, as in WGANs [15], that encourages the representations of the different datasets to be as close as possible. Unlike ADDA [19], the emotion classifier and database critic are iteratively trained, ensuring the presence of emotion in the intermediate representation. We hypothesize that this creates a more generalized representation of emotion that will perform better

**Algorithm 1** Train ADDoG for one epoch. We use the default values of $n_{critic} = 5, c = 0.01, m = 32, \alpha = 0.0001, \beta_1 = 0.9, \beta_2 = 0.999$

**Require:** The number of critic iterations per generator/classifier iteration $n_{critic}$, the critic clipping range $c$, the batch size $m$, Adam hyperparameters $\alpha, \beta_1, \beta_2$.

**Require:** Generator parameters $\phi$, critic parameters $\psi$, emotion classifier parameters $\theta$.

**Require:** Emotion class weights for SRC $S_w$, emotion class weights for labelled TAR $L_w$

1: $n \leftarrow$ (Number of SRC samples) / $m$
2: **for** $batch = 1,..., n$ **do**
3:     **for** $t = 1,..., n_{critic}$ **do**
4:         Sample $\{S_X^{(i)}\}_{i=1}^m$ a batch from SRC data
5:         Sample $\{T_X^{(i)}\}_{i=1}^m$ a batch from TAR data
6:         $S_R \leftarrow G_\phi(S_X)$            ▷ Encoded SRC
7:         $T_R \leftarrow G_\phi(T_X)$            ▷ Encoded TAR
8:         $loss \leftarrow \frac{1}{m}\sum_{i=1}^m C_\psi(S_R^{(i)}) - \frac{1}{m}\sum_{i=1}^m C_\psi(T_R^{(i)})$
9:         $\psi \leftarrow$ Adam$(\Delta_\psi[loss], \psi, \alpha, \beta_1, \beta_2)$
10:         $\psi \leftarrow$ clip$(\psi, -c, c)$     ▷ Clip critic weights
11:     **end for**
12:     Sample $\{S_X^{(i)}, S_y^{(i)}\}_{i=1}^m$ a batch from SRC data
13:     Sample $\{T_X^{(i)}\}_{i=1}^m$ a batch from **all** TAR
14:     Sample $\{L_X^{(i)}, L_y^{(i)}\}_{i=1}^m$ a batch from **labelled** TAR
15:     $S_R \leftarrow G_\phi(S_X)$            ▷ Encoded SRC
16:     $T_R \leftarrow G_\phi(T_X)$            ▷ Encoded TAR
17:     $L_R \leftarrow G_\phi(L_X)$        ▷ Encoded labelled TAR
18:     $loss_C \leftarrow \frac{1}{m}\sum_{i=1}^m C_\psi(T_R^{(i)}) - \frac{1}{m}\sum_{i=1}^m C_\psi(S_R^{(i)})$
19:     $loss_E \leftarrow -\frac{1}{m}\sum_{i=1}^m S_y^{(i)} \times log(E_\theta(S_R^{(i)})) \times S_w$
          $-\frac{1}{m}\sum_{i=1}^m L_y^{(i)} \times log(E_\theta(L_R^{(i)})) \times L_w$
20:     $\phi \leftarrow$ Adam$(\Delta_\phi[loss_C + loss_E], \phi, \alpha, \beta_1, \beta_2)$
21:     $\theta \leftarrow$ Adam$(\Delta_\theta[loss_E], \theta, \alpha, \beta_1, \beta_2)$
22: **end for**

across datasets, compared with CNN. This is because the representation will remove unrelated information that could mislead the emotion classifier (environment noise, microphone quality, subject demographics). Figure 3 shows the network structure of ADDoG. It consists of three main components: (1) the feature encoder (convolutions + max pooling); (2) the emotion classifier (fully connected layers + softmax); (3) the critic (fully connected layers + linear output).

The ADDoG hyperparameters are identical to those used in CNN. The critic network follows the same structure and hyperparameters as the emotion classifier. The only difference is that the critic is a linear activation instead of a softmax layer [15]. The training of the ADDoG network follows Algorithm 1 during each epoch. The number of training iterations per epoch is equal to the number of utterances in SRC divided by the batch size. Each iteration is divided into two main phases: (1) training the critic; (2) training the feature encoder and emotion classifier.

**Training the critic:** We freeze the weights in the feature encoder and emotion classifier. First, unlabelled batches are sampled from SRC and TAR. Next, the MFBs are passed through the feature encoder to get the intermediate representations. These intermediate representations are then passed to the critic. We calculate the loss by subtracting the mean TAR output from the mean SRC output. We use the Adam optimizer on the critic weights with this loss to encourage TAR outputs to be as large as possible and SRC outputs to be as small as possible, estimating the Wasserstein, or earth mover's, distance [15]. The critic weights are then clipped to a range between -0.01 and 0.01, as in [15], to keep the outputs from growing infinitely. This critic training process is repeated five times to fully converge the critic before training the other systems, as in the original WGAN paper.

**Training the feature encoder and emotion classifier:** We freeze the critic weights. Next, we sample batches from SRC, TAR, and the subset of TAR that is labelled (if any). We then pass the MFBs through the entire network, getting outputs from the emotion classifier and the critic. As in the CNN training method, we calculate the emotion loss by weighted cross entropy using the SRC and labelled TAR sets. We add an additional term to the loss function for the critic that aims to move the dataset representations closer to one another. This is calculated by subtracting the mean SRC output from the mean TAR output, inverting the Wasserstein distance.

This training procedure iteratively moves the two dataset representations closer to one another, while following a clear gradient at each step. In contrast, DANNs [20] attempt to make a more generalized representation by "unlearning" domain. We attempted to implement DANN for our preliminary cross-corpus experiments, but had issues with getting the results to converge, as alluded to in prior work [21]. Our method of ADDoG gets around this by having a clearly defined target at each step to "meet in the middle" and also utilizes the Wasserstein distance instead of a traditional discriminator.

### 4.3 MADDoG

Multiclass Adversarial Discriminative Domain Generalization (MADDoG) expands the ADDoG algorithm to allow for more than two datasets. The MADDoG network structure is identical to ADDoG (Figure 3) except the critic has an output for each dataset instead of a single output. This allows for the method to account for the differences between all datasets while learning the representation. In contrast, ADDoG requires datasets to be grouped into target and source sets, not considering the differences within the sets.

The training of the MADDoG network follows Algorithm 2 during each epoch. The number of training iterations per epoch is equal to the number of utterances in SRC and TAR divided by the batch size. This is because data are drawn from all datasets simultaneously when training the critic instead of each separately. As in ADDoG, each iteration is divided into two main phases: (1) training the critic; (2) training the feature encoder and emotion classifier.

**Training the critic:** Training the critic is similar to ADDoG and begins with freezing the weights in the feature encoder and emotion classifier. One unlabelled batch is sampled from the combined SRC and TAR sets. The MFBs are then passed through the network to get the critic outputs, which are then modified as follows:

1) We calculate the proportion of each dataset versus the occurrence of all other datasets.

2) For each utterance in the batch, we flip the critic output corresponding to its dataset and multiply it by the previously calculated one-versus-all weight.

**Algorithm 2** Train MADDoG for one epoch. We use the default values of $n_{critic} = 5$, $c = 0.01$, $m = 32$, $\alpha = 0.0001$, $\beta_1 = 0.9$, $\beta_2 = 0.999$, $\lambda = 0.1$

**Require:** The number of critic iterations per generator/classifier iteration $n_{critic}$, the critic clipping range $c$, the batch size $m$, Adam hyperparameters $\alpha, \beta_1, \beta_2$, the dataset generalization parameter $\lambda$.

**Require:** Generator parameters $\phi$, critic parameters $\psi$, emotion classifier parameters $\theta$.

**Require:** Emotion class weights for SRC $S_w$, emotion class weights for labelled TAR $L_w$, dataset one-versus-all weights $DS_w$

1: $n \leftarrow$ (Number of SRC and TAR samples) / $m$
2: **for** $batch = 1, ..., n$ **do**
3:      **for** $t = 1, ..., n_{critic}$ **do**
4:         Sample $\{X^{(i)}, ds^{(i)}\}_{i=1}^m$ a batch from all data
5:         $R \leftarrow G_\phi(X)$                $\triangleright$ Encoded data
6:         $D \leftarrow C_\psi(R)$        $\triangleright$ Get the 3 outputs of critic
7:         $D^{(:,ds)} \leftarrow D^{(:,ds)} \times -DS_w^{(ds)}$
8:         $loss \leftarrow -\frac{1}{m} \frac{1}{3} \sum_{i=1}^m \sum_{j=1}^3 D^{(i,j)}$
9:         $\psi \leftarrow \text{Adam}(\Delta_\psi[loss], \psi, \alpha, \beta_1, \beta_2)$
10:        $\psi \leftarrow \text{clip}(\psi, -c, c)$         $\triangleright$ Clip critic weights
11:      **end for**
12:      Sample $\{S_X^{(i)}, S_y^{(i)}, S_{ds}^{(i)}\}_{i=1}^m$ a batch from SRC data
13:      Sample $\{T_X^{(i)}, T_{ds}^{(i)}\}_{i=1}^m$ a batch from **all** TAR
14:      Sample $\{L_X^{(i)}, L_y^{(i)}\}_{i=1}^m$ a batch from **labelled** TAR
15:      $S_R \leftarrow G_\phi(S_X)$               $\triangleright$ Encoded SRC
16:      $T_R \leftarrow G_\phi(T_X)$               $\triangleright$ Encoded TAR
17:      $L_R \leftarrow G_\phi(L_X)$        $\triangleright$ Encoded labelled TAR
18:      $loss_C \leftarrow \frac{1}{m} \sum_{i=1}^m C_\psi(T_R^{(i)}) \times T_{ds}^{(i)}$
                 $+ \frac{1}{m} \sum_{i=1}^m C_\psi(S_R^{(i)}) \times S_{ds}^{(i)}$
19:      $loss_E \leftarrow -\frac{1}{m} \sum_{i=1}^m S_y^{(i)} \times log(E_\theta(S_R^{(i)})) \times S_w$
                 $-\frac{1}{m} \sum_{i=1}^m L_y^{(i)} \times log(E_\theta(L_R^{(i)})) \times L_w$
20:      $\phi \leftarrow \text{Adam}(\Delta_\phi[\lambda \times loss_C + loss_E], \phi, \alpha, \beta_1, \beta_2)$
21:      $\theta \leftarrow \text{Adam}(\Delta_\theta[loss_E], \theta, \alpha, \beta_1, \beta_2)$
22: **end for**

This makes each critic output a one-versus-all dataset critic and weights each of them so that samples from inside and outside the dataset are given equal total weight. The critic loss is calculated as the mean of the critic outputs, causing all of them to trend smaller. However, because the within dataset output is flipped, it is encouraged to be larger. The critic loss estimates the one-versus-all Wasserstein distance for each dataset. As in ADDoG, the critic weights are clipped between -0.01 and 0.01 and the whole process is repeated five times.

**Training the feature encoder and emotion classifier:** The critic weights are first frozen. We then sample batches from SRC, TAR, and the subset of TAR that is labelled (if any). The MFBs are passed through the network, providing the emotion classifier and critic outputs. We calculate the emotion loss as before, using weighted cross entropy over the SRC and labelled TAR data. For each utterance, the contribution to the critic loss is the critic output from the same dataset as the utterance (ignoring the other outputs). This encourages the one-versus-all Wasserstein distance to be reduced and the dataset to start looking like the others. Because this results in a more complex learning procedure than before, in practice we need to provide a weighting parameter $\lambda = 0.1$ (found in preliminary experiments) to allow for the representation to converge. The total loss is the emotion loss added to the critic loss times $\lambda$.

The novelty of the MADDoG method is its ability to incorporate multiple datasets into its generalization procedure while still maintaining a clearly defined target (the other datasets) at each step. This creates cross-dataset representations that become more similar as the system is trained and that continue to encode the emotion information in the signal. As long as a sufficiently small learning rate is used, the intermediate dataset representations should converge somewhere in the middle. If instead, SRC datasets were considered as a group, the training procedure would not do anything to generalize between the SRC datasets. MADDoG considers these differences to enforce a more generalized representation that allows for better cross-corpus performance.

## 5 EXPERIMENTAL DESIGN

We design four sets of experiments to examine different types of cross-dataset emotion classification. Each experiment examines the effect of the inclusion or absence of labelled data in the target dataset. The final two experiments focus on incorporating both laboratory and in-the-wild datasets.

All experiments begin by dividing the data into three folds: train, validation, and test. We run each experiment for 30 epochs, recording validation performance and test set predictions at each step. In this paper, we use Unweighted Average Recall (UAR) as the performance metric. This ensures that each valence class is given equal weight, despite possible imbalance, and results in a chance performance of 0.33 UAR. Once an experiment is complete, we record test predictions from the highest validation epoch to prevent overfitting and calculate the UAR for each test subject. Each experiment is repeated a total ten times (fifty times for Experiment 1, Section 5.1), resulting in a final performance matrix of size (Number of Repeats × Number of Subjects). Folds are kept consistent between different methods so that these performance matrices can be compared. These experiments were split between two machines with GPUs - one with four GeForce GTX 1080s and another with one GeForce GTX 1080 and two Titan X's.

### 5.1 Experiment 1: Cross-Dataset

We determine the effect of training and testing on different datasets when labelled data is unavailable in the target dataset. Additionally, we constrain this initial experiment to only consider data from similar environments - using IEMOCAP and MSP-Improv, which were both recorded in a laboratory. We form the train and validation sets by splitting the SRC data randomly on a 80:20 split, respectively. In this experiment, we compare the effectiveness of a CNN versus ADDoG, which uses the unlabelled test data in the training process to learn a more generalized representation for emotion. We run each experiment for 50 total repeats, so that we get enough per-epoch data to perform a convergence analysis.

### 5.2 Experiment 2: Increasing Target Labels

The next experiment augments the training data with varying amounts of labelled examples from TAR. This experiment continues focusing on the laboratory recorded datasets IEMOCAP and MSP-improv. We train the network with 0, 200, 400, 800, 1,600, and 3,200 labelled TAR utterances, in addition to the labelled SRC utterances and unlabelled test utterances.

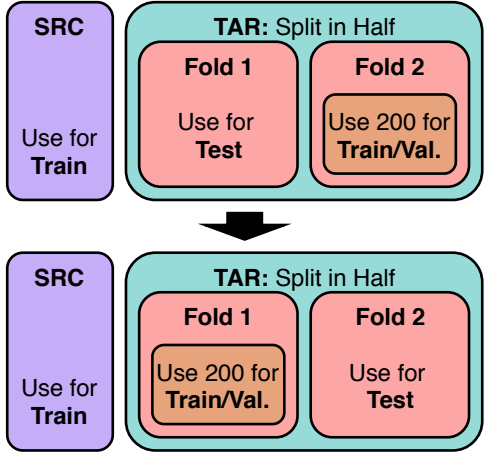

Fig. 4. Folds used for Experiments 2, 3, and 4 when 200 labelled TAR are available. The SRC set is always used as part of the train set. The TAR set is split in half - part for testing and part for randomly sampling the 200 labelled TAR. The TAR data not selected in Fold 2 is discarded. After getting test predictions, the TAR folds are swapped and the process is repeated.

We follow the fold scheme shown in Figure 4 to get test predictions for all utterances in TAR. The number of SRC utterances and test utterances is kept constant through all experiments. SRC utterances are only used in the train set, unless no labelled TAR data is available. In that case, the SRC data follows a random 80:20 split between train and validation, as in Experiment 1. We split the TAR data randomly in half to allow for some labelled data for training, while reserving the other half for testing. If labelled data is used for an experiment, these samples are drawn from one of the halves and split 80:20 between the train and validation sets. Figure 4 depicts the case of having 200 labelled TAR utterances, resulting in a validation set of 40 labelled TAR and a train set including 160 labelled TAR and all of SRC. The remaining TAR data in the fold is discarded so the amount of unlabelled data is kept constant. This procedure results in test predictions for half of TAR. The TAR folds are then swapped, a new model is trained, and test predictions on the other half are output. Finally, we calculate the UARs for each subject in TAR using the concatenated predictions.

ADDoG is able to use the unlabelled test data along with the labelled SRC and TAR data during training for learning a more generalized dataset representation. The baseline CNN method is provided labelled data from both SRC and TAR when available. We also introduce another baseline method that specializes (SP) on the available labelled target data. SP uses the same network and training procedure as CNN, but only uses labelled TAR. Because of this, it is unable to be trained when 0 labelled TAR utterances are provided. We run this an all other experiments using 10 total repeats.

### 5.3 Experiment 3: To In-the-Wild Data

We next examine the effect of training on a laboratory recorded dataset (IEMOCAP and/or MSP-Improv) and testing on emotion in-the-wild (PRIORI Emotion). We expect this experiment to be more difficult than the previous two, due to the difference

| | MSP-Improv to IEMOCAP | IEMOCAP to MSP-Improv |
|---|---|---|
| CNN | 0.439±0.022 | 0.432±0.012 |
| ADDoG | **0.474±0.009*** | **0.444±0.007*** |

in recording environment (combining lab and cellphone call), recording quality (previously 16 kHz, now 8 kHz), and elicitation strategy (combining acted and natural conversation). We examine the effect of training on IEMOCAP or MSP-Improv alone, as well as training on them together. Each test follows the same procedure as Experiment 2, using the folds seen in Figure 4. We again compare the CNN, SP, and ADDoG methods. The experiment combining IEMOCAP and MSP-Improv training data also employs the MADDoG method to take advantage of all three datasets.

### 5.4 Experiment 4: From In-the-Wild Data

Our final experiment examines the reverse of Experiment 3 - training on in-the-wild data (PRIORI Emotion) and testing on laboratory recorded emotion (IEMOCAP or MSP-Improv). The experiment follows the same strategy as Experiment 2, using the fold scheme seen in Figure 4. We use the CNN, SP, and ADDoG models. Unlike Experiment 3, MADDoG is not used, as PRIORI Emotion is the only dataset used for training.

## 6 RESULTS

In all of the presented UARs, errors are calculated by first taking the mean subject UAR within each repeat of an experiment. The reported errors are the standard deviation of these means across all repeats, showing the stability of the findings. Matplotlib [55] was used to generate all result plots with the error shown as shaded error bands. Significance is determined using an analysis of variance in R [56] over the matrix of subject UARs output by the compared methods, as explained in Section 5. Significant results in each experiment are indicated by dots on the plots and/or bolded and starred values in the tables.

### 6.1 Experiment 1: Cross-Dataset

Table 2 shows the results of using IEMOCAP as SRC and MSP-Improv as TAR, as well as the reverse experiment. Two different methods are compared, including the CNN, which is trained only using the SRC data, and ADDoG, which is additionally trained using the unlabelled TAR data, creating a more generalized intermediate representation. We find that ADDoG significantly outperforms CNN in both cases, implying that a more generalized representation can be used to improve cross-corpus testing without added labelled data.

In addition, we note that the standard deviation across experiment repeats is much lower for ADDoG versus CNN. This can also be seen in the convergence of results, as seen in Figure 5. Figure 5a in particular shows ADDA with a much smaller error at each epoch, compared with CNN. This is especially important for cross-corpus testing where labelled data is not available in the target dataset for validation. Using SRC data for validation is necessary in these experiments, but can still be unreliable, due to the mismatch. This may be less of a problem for ADDoG because of the results stability, contributing to the overall better performance.

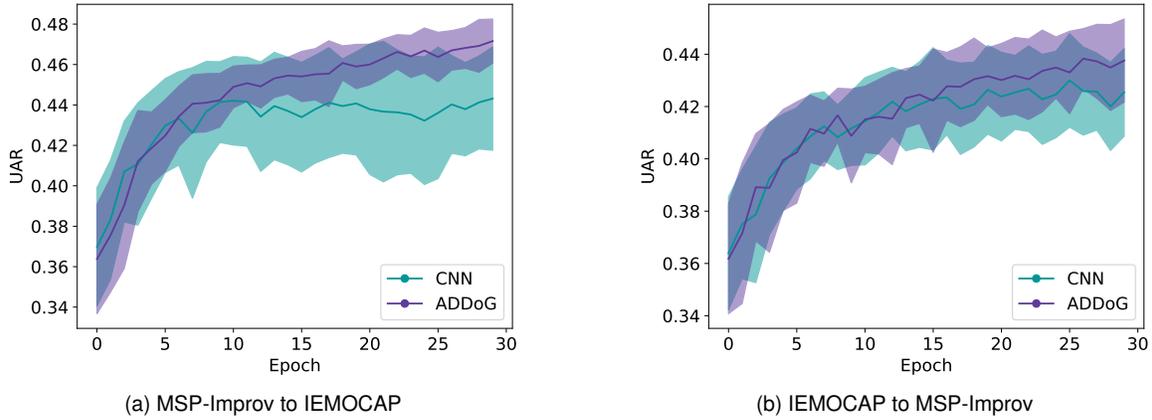

Fig. 5. The test set mean subject UAR at different epochs when training on one dataset and testing on another. In particular, Figure 5a demonstrates how ADDoG reduces the variance of the output, improving cross-corpus testing, regardless of the mismatched validation set.

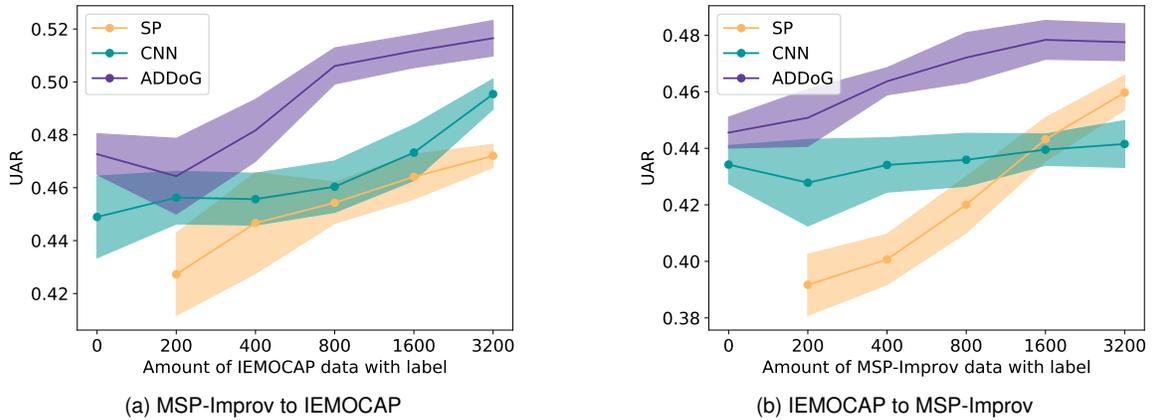

Fig. 6. Results of training on either IEMOCAP or MSP-Improv and testing on the other with increasing amounts of labels from the target dataset. Dots indicate methods significantly different from ADDoG using an analysis of variance in R (p=0.05).

## 6.2 Experiment 2: Increasing Target Labels

Figure 6 shows the results for Experiment 2, when we begin to incorporate labelled TAR data into the training and validation methodology. The left most point on both plots is the case when only unlabelled TAR data is available. This is slightly different than Experiment 1, as only half the amount of unlabelled data is available due to the fold structure shown in Figure 4. We find that ADDoG significantly improves on the baseline method in all cases, although the margin of improvement decreases with larger amounts of labelled target data. This may indicate that generalizing the representation may have diminishing returns once there is sufficient labelled data in the target domain. However, coupling even a small amount of labelled data and ADDoG results in significant improvement over baseline methods.

Adding labelled data to the ADDoG method increases its performance in all but one case - training on MSP-Improv and testing on IEMOCAP with only 200 labelled IEMOCAP utterances. While still significantly better than CNN and SP with the same amount of labelled data, better performance is actually attained using ADDoG without labelled IEMOCAP data. This could be due to the relatively small validation set, only consisting of 20% of the labelled data, or 40 utterances.

This may not provide a reliable enough estimate of test performance, resulting in the larger error band around the result. It may be better to instead incorporate some additional SRC data in validation when very small amounts of TAR data are only available.

We also find that the SP method begins to outperform the CNN method when training on IEMOCAP and testing on MSP-Improv once a large amount of MSP-Improv labelled data is included. This may imply that appending SRC data to TAR data only complicates the training when not considering the effect of dataset. This is even more apparent when considering very different datasets, as seen in the next section.

## 6.3 Experiment 3: To In-the-Wild Data

The results of Experiment 3 are shown Figure 7. Experiment 3 considers the effect of training on a laboratory recorded data (IEMOCAP and/or MSP-Improv) and testing on an in-the-wild set (PRIORI Emotion). Because all experiments use the same test set, the y-axis (UAR) range is kept constant. The SP results are the same between all figures, as it does not rely on the SRC data.

The first two figures 7a and 7b examine the case where just one laboratory dataset is used to train the model. We

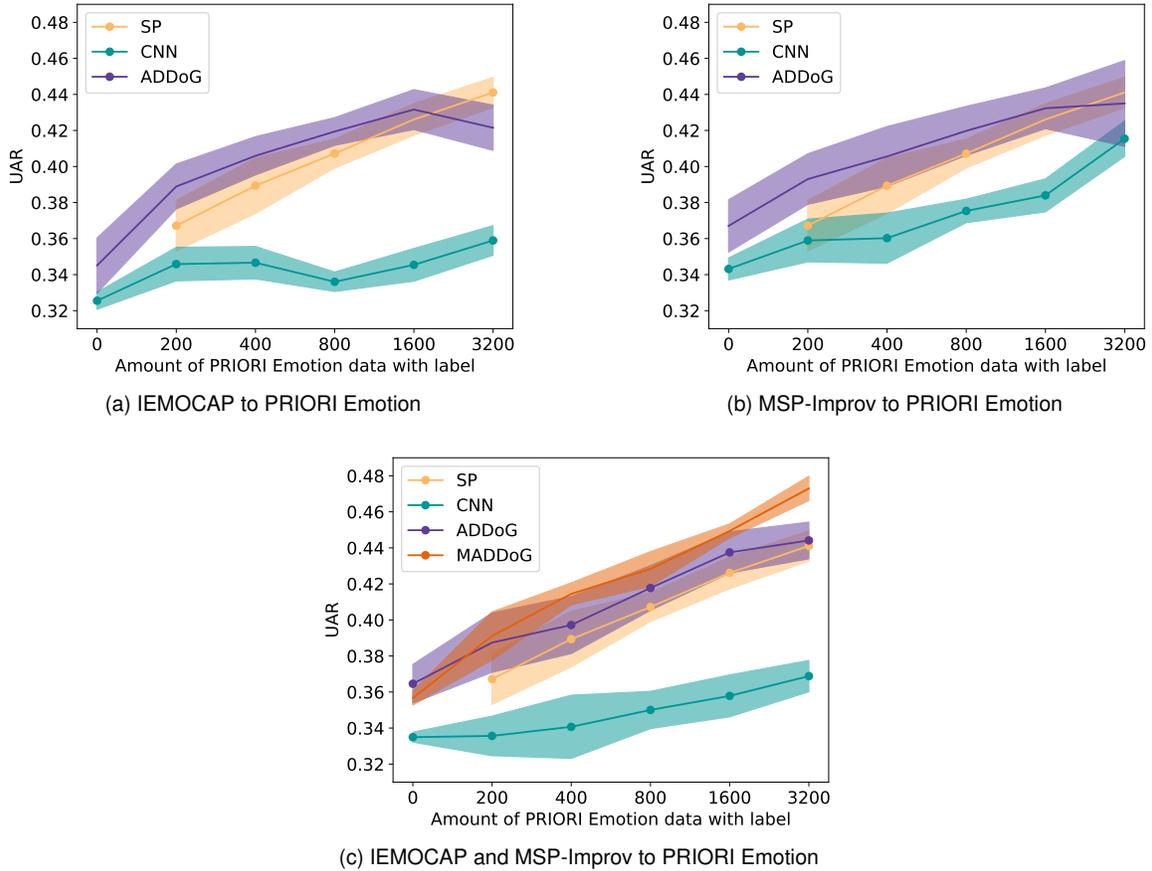

(a) IEMOCAP to PRIORI Emotion

(b) MSP-Improv to PRIORI Emotion

(c) IEMOCAP and MSP-Improv to PRIORI Emotion

Fig. 7. Results of training on IEMOCAP and/or MSP-Improv and testing on PRIORI Emotion with increasing amounts of labels from PRIORI Emotion. Dots indicate methods significantly different from ADDoG in (a) and (b) and MADDoG in (c) using an analysis of variance in R (p=0.05).

find much lower performance than prior experiments, due to the mismatch in recording conditions and elicitation strategy (acted versus a natural phone conversation). Combining SRC and TAR data together with the CNN method entirely fails, with the results consistently being the worst, due to the extreme mismatch. However, ADDoG is still able to provide a significant improvement in performance when none or a small amount (800 or fewer) of labelled TAR utterances are available. For these experiments with smaller labelled TAR data, the advantage of using ADDoG is approximately similar to that attained by doubling the amount of labelled TAR data. However, this trend is broken with larger amount of labelled TAR data where AD-DoG no longer performs better and is in one case significantly worse (IEMOCAP to PRIORI, 3200 labelled samples). Due to the mismatch in dataset, it is better to specialize a model to the dataset characteristics, instead of generalizing, once a certain critical mass is attained. Both CNN and ADDoG perform slightly better when trained with MSP-Improv data, implying that it may be the more similar of the two datasets to PRIORI Emotion. This could potentially be due to the included more natural speech in the MSP-Improv dataset recorded in between scenarios.

Figure 7c shows the results for the last case where IEMOCAP and MSP-Improv are both simultaneously considered as SRC datasets. Despite the added data, the CNN method is unable to perform better than with just MSP-Improv data, implying that the additional dataset is just confusing the classifier. The

ADDoG classifier is able to take advantage of the additional data to at least perform the same as, if not better than, the MSP-Improv ADDoG method. While the method is not hurt by the addition of IEMOCAP, in most cases it does not help. However, MADDoG performs better than all methods using labelled data (significantly in all cases but ADDoG with 200 samples). This is likely due to the fact that it is able to effectively integrate together information from all datasets and come up with an even more generalized representation. ADDoG still seems to perform significantly better in the case where there is no labelled TAR data. Perhaps the labels from the other two datasets dominate the representation when none are available for MADDoG.

### 6.4 Experiment 4: From In-the-Wild Data

Because of our success in generalizing a representation across laboratory and in-the-wild datasets, we were interested in cross-corpus testing in the reverse direction. Figure 8 shows the results when training on PRIORI Emotion and testing on either IEMOCAP or MSP-Improv. In these experiments we just use the CNN, SP, and ADDoG methods, as there is only one SRC dataset included, making MADDoG unnecessary. Our results again show that the CNN method performs the worst, demonstrating that appending together datasets does not work effectively when the datasets are too different. ADDoG also behaves similarly to Experiment 3 with significant improvements in most cases without labelled data or small amounts of labelled data. SP has similar or better performance than ADDoG once a substantial amount of labelled TAR data is available, implying that a method that trades off between generalization and specialization may instead be needed in these cases.

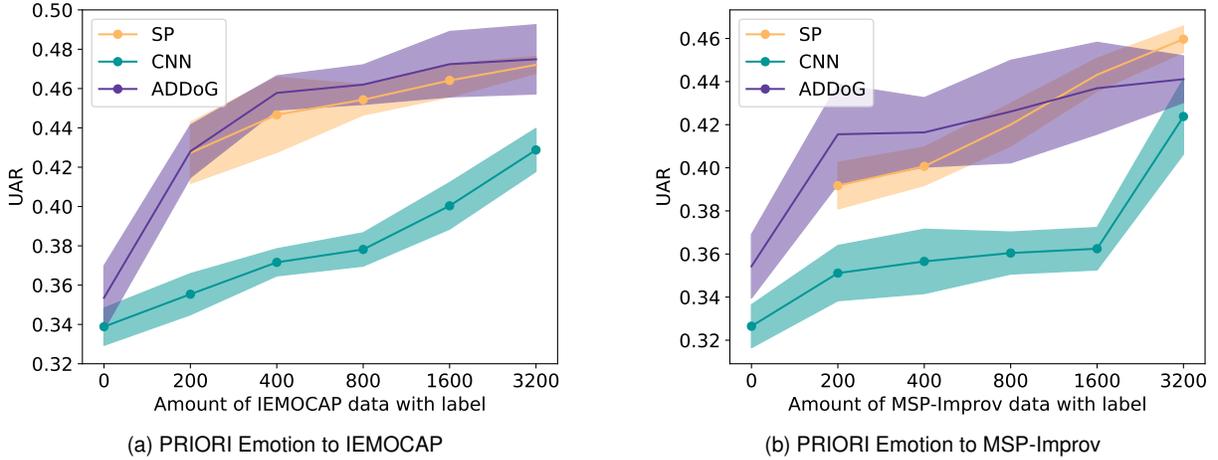

(a) PRIORI Emotion to IEMOCAP

(b) PRIORI Emotion to MSP-Improv

Fig. 8. Results of training on PRIORI Emotion and testing on another dataset with increasing amounts of labels from the target dataset. Dots indicate methods significantly different from ADDoG using an analysis of variance in R (p=0.05).

## 7 DISCUSSION AND CONCLUSION

In this paper, we investigate methods of controlling for the unwanted factors of variation when conducting cross-corpus experiments. These factors can include environmental noise, recording device differences, elicitation strategies (acted versus natural), and subject demographics. In cross-corpus speech emotion recognition, these factors can distract from the underlying emotion and decrease performance, especially because of the often smaller dataset sizes.

We introduce two new methods, ADDoG and MADDoG, which aim to generalize the representation of speech emotion across datasets. Both methods iteratively move their dataset representations closer to one another and have a clearly defined target at each step, following a "meet in the middle" approach. Experiments 1 and 2 focus on more traditional laboratory datasets to introduce the models and explore convergence. Experiments 3 and 4 take advantage of the PRIORI Emotion dataset to examine the effect of training with in-the-wild data. Experiment 3 also explores training with three simultaneous datasets using the MADDoG method.

Our results indicate that ADDoG is able to consistently converge and produce a more generalized representation across datasets, even when no labelled target data is available. Significant improvement is found with no added labelled data in all four experiments, regardless of the number of datasets or whether they are laboratory or in-the-wild recordings. These results reinforce the idea that the "meeting in the middle" approach of ADDoG can reach the same generalized representation as "unlearning", seen with DANNs [20], [21]. However, convergence of the algorithm is easier to attain because of the more straightforward training paradigm. This generalized representation not only improves performance, but also decreases variance over different repeats of the experiment with different data. Because of this stability, less emphasis needs to be placed on validation. This is particularly important, since the validation and test sets are mismatched when conducting cross-corpus experiments.

Further experiments demonstrate how to effectively use small amounts of target labelled data when available. Simply combining the labelled data together from both datasets performs reasonably well when the recording conditions closely match, such as those in the two laboratory datasets - IEMOCAP and MSP-Improv. However, this method fails when substantially different data is introduced, such as PRIORI Emotion, demonstrated by the low CNN results in Experiments 3 and 4. ADDoG takes a more elegant approach to combining these datasets by building a generalized model and ensuring the representation is valid for the provided TAR data. Additionally, for the case of more than two datasets, MADDoG is able to recognize the differentiating factors in all SRC and take advantage of them. While ADDoG and MADDoG are significantly the best methods in most cases, SP performs comparably or better when enough TAR labels are available and TAR and SRC are very different. This indicates that generalized models can only go so far and it is important to understand when a domain is so different that a specialized model may be needed.

Future experiments will explore other factors of speech variation besides dataset - including gender, phoneme, subject, and recording device. ADDoG and MADDoG should be able to be used to find a representation for emotion that is more consistent across these factors. For example, while conducting traditional leave-one-subject-out speech emotion testing, it would be possible to consider the training subjects as SRC and the test subject as TAR. We hypothesize that in this case it would be preferred to find a more generalized representation of emotion that removes the contribution of subject characteristics. Similarly, we could use the technique to reduce the impact of recordings taken over multiple devices, previously shown to reduce mood classification performance [5].

Additionally, we are interested in exploring the trade-off between building systems specialized for certain domains versus building generalized representations. Once enough labelled data is available in a target domain, it is often better to just specialize with it instead of using information from other domains, as seen in Section 6.3. Because of this, it is important to formulate exactly when to generalize and when to specialize.

We plan on investigating a method that fuses both approaches and estimates which is better, given that domain data currently present.

Finally, we plan to use the techniques explored in this paper to facilitate the estimation of mood in-the-wild. As explained in Section 3.3, the PRIORI Emotion dataset is part of the larger PRIORI dataset, which captures the everyday speech of individuals with bipolar disorder. We aim to use predicted emotion as a mid-level feature for mood estimation, as first explored in [41]. By employing the techniques of ADDoG and MADDoG, we will bring together multiple emotion and mood datasets to build a model capable of working effectively in-the-wild.

## ACKNOWLEDGMENTS

This work was supported by the National Science Foundation (CAREER-1651740). National Institute of Mental Health (R01MH108610, R34MH100404) , the Heinz C Prechter Bipolar Research Fund, and the Richard Tam Foundation at the University of Michigan.

## REFERENCES

[1] R. Beale and C. Peter, "The role of affect and emotion in hci," in *Affect and emotion in human-computer interaction.* Springer, 2008, pp. 1–11.

[2] C.-N. Anagnostopoulos, T. Iliou, and I. Giannoukos, "Features and classifiers for emotion recognition from speech: a survey from 2000 to 2011," *Artificial Intelligence Review*, vol. 43, no. 2, pp. 155–177, 2015.

[3] B. Schuller, B. Vlasenko, F. Eyben, M. Wollmer, A. Stuhlsatz, A. Wendemuth, and G. Rigoll, "Cross-corpus acoustic emotion recognition: Variances and strategies," *IEEE Transactions on Affective Computing*, vol. 1, no. 2, pp. 119–131, 2010.

[4] J. Deng, Z. Zhang, F. Eyben, and B. Schuller, "Autoencoder-based unsupervised domain adaptation for speech emotion recognition," *IEEE Signal Processing Letters*, vol. 21, no. 9, pp. 1068–1072, 2014.

[5] J. Gideon, E. M. Provost, and M. McInnis, "Mood state prediction from speech of varying acoustic quality for individuals with bipolar disorder," in *Acoustics, Speech and Signal Processing (ICASSP), 2016 IEEE International Conference on.* IEEE, 2016, pp. 2359–2363.

[6] J. Kim, G. Englebienne, K. P. Truong, and V. Evers, "Towards speech emotion recognition" in the wild" using aggregated corpora and deep multi-task learning," *arXiv preprint arXiv:1708.03920*, 2017.

[7] Z. Zhang, F. Weninger, M. Wöllmer, and B. Schuller, "Unsupervised learning in cross-corpus acoustic emotion recognition," in *Automatic Speech Recognition and Understanding (ASRU), 2011 IEEE Workshop on.* IEEE, 2011, pp. 523–528.

[8] B. Schuller, Z. Zhang, F. Weninger, and G. Rigoll, "Selecting training data for cross-corpus speech emotion recognition: Prototypicality vs. generalization," in *Proc. Afeka-AVIOS Speech Processing Conference, Tel Aviv, Israel.* Citeseer, 2011.

[9] ——, "Using multiple databases for training in emotion recognition: To unite or to vote?" in *Twelfth Annual Conference of the International Speech Communication Association*, 2011.

[10] S. Parthasarathy and C. Busso, "Jointly predicting arousal, valence and dominance with multi-task learning," *INTERSPEECH, Stockholm, Sweden*, 2017.

[11] I. Goodfellow, J. Pouget-Abadie, M. Mirza, B. Xu, D. Warde-Farley, S. Ozair, A. Courville, and Y. Bengio, "Generative adversarial nets," in *Advances in neural information processing systems*, 2014, pp. 2672–2680.

[12] A. Radford, L. Metz, and S. Chintala, "Unsupervised representation learning with deep convolutional generative adversarial networks," *arXiv preprint arXiv:1511.06434*, 2015.

[13] S. Sahu, R. Gupta, and C. Espy-Wilson, "On enhancing speech emotion recognition using generative adversarial networks," *arXiv preprint arXiv:1806.06626*, 2018.

[14] J. Chang and S. Scherer, "Learning representations of emotional speech with deep convolutional generative adversarial networks," in *Acoustics, Speech and Signal Processing (ICASSP), 2017 IEEE International Conference on.* IEEE, 2017, pp. 2746–2750.

[15] M. Arjovsky, S. Chintala, and L. Bottou, "Wasserstein gan," *arXiv preprint arXiv:1701.07875*, 2017.

[16] J.-Y. Zhu, T. Park, P. Isola, and A. A. Efros, "Unpaired image-to-image translation using cycle-consistent adversarial networks," *arXiv preprint*, 2017.

[17] X. Zhu, Y. Liu, J. Li, T. Wan, and Z. Qin, "Emotion classification with data augmentation using generative adversarial networks," in *Pacific-Asia Conference on Knowledge Discovery and Data Mining.* Springer, 2018, pp. 349–360.

[18] T. Kaneko and H. Kameoka, "Parallel-data-free voice conversion using cycle-consistent adversarial networks," *arXiv preprint arXiv:1711.11293*, 2017.

[19] E. Tzeng, J. Hoffman, K. Saenko, and T. Darrell, "Adversarial discriminative domain adaptation," in *Computer Vision and Pattern Recognition (CVPR)*, vol. 1, no. 2, 2017, p. 4.

[20] H. Ajakan, P. Germain, H. Larochelle, F. Laviolette, and M. Marchand, "Domain-adversarial neural networks," *arXiv preprint arXiv:1412.4446*, 2014.

[21] M. Abdelwahab and C. Busso, "Domain adversarial for acoustic emotion recognition," *arXiv preprint arXiv:1804.07690*, 2018.

[22] Z. Huang, M. Dong, Q. Mao, and Y. Zhan, "Speech emotion recognition using cnn," in *Proceedings of the 22nd ACM international conference on Multimedia.* ACM, 2014, pp. 801–804.

[23] Z. Aldeneh and E. M. Provost, "Using regional saliency for speech emotion recognition," in *Acoustics, Speech and Signal Processing (ICASSP), 2017 IEEE International Conference on.* IEEE, 2017, pp. 2741–2745.

[24] B. Zhang, G. Essl, and E. Mower Provost, "Predicting the distribution of emotion perception: capturing inter-rater variability," in *Proceedings of the 19th ACM International Conference on Multimodal Interaction.* ACM, 2017, pp. 51–59.

[25] Y. Zong, W. Zheng, X. Huang, K. Yan, J. Yan, and T. Zhang, "Emotion recognition in the wild via sparse transductive transfer linear discriminant analysis," *Journal on Multimodal User Interfaces*, vol. 10, no. 2, pp. 163–172, 2016.

[26] P. Song, "Transfer linear subspace learning for cross-corpus speech emotion recognition," *IEEE Transactions on Affective Computing*, 2017.

[27] J. Deng, Z. Zhang, and B. Schuller, "Linked source and target domain subspace feature transfer learning–exemplified by speech emotion recognition," in *Pattern Recognition (ICPR), 2014 22nd International Conference on.* IEEE, 2014, pp. 761–766.

[28] S. Latif, R. Rana, J. Qadir, and J. Epps, "Variational autoencoders for learning latent representations of speech emotion," *arXiv preprint arXiv:1712.08708*, 2017.

[29] S. E. Eskimez, Z. Duan, and W. Heinzelman, "Unsupervised learning approach to feature analysis for automatic speech emotion recognition," 2018.

[30] S. Sahu, R. Gupta, G. Sivaraman, W. AbdAlmageed, and C. Espy-Wilson, "Adversarial auto-encoders for speech based emotion recognition," *arXiv preprint arXiv:1806.02146*, 2018.

[31] H. Li, S. J. Pan, S. Wang, and A. C. Kot, "Domain generalization with adversarial feature learning," in *Proc. IEEE Conf. Comput. Vis. Pattern Recognit.(CVPR)*, 2018.

[32] A. Hassan, R. Damper, and M. Niranjan, "On acoustic emotion recognition: compensating for covariate shift," *IEEE Transactions on Audio, Speech, and Language Processing*, vol. 21, no. 7, pp. 1458–1468, 2013.

[33] P. Song, Y. Jin, L. Zhao, and M. Xin, "Speech emotion recognition using transfer learning," *IEICE TRANSACTIONS on Information and Systems*, vol. 97, no. 9, pp. 2530–2532, 2014.

[34] X. He and P. Niyogi, "Locality preserving projections," in *Advances in neural information processing systems*, 2004, pp. 153–160.

[35] M. Abdelwahab and C. Busso, "Supervised domain adaptation for emotion recognition from speech," in *Acoustics, Speech and Signal Processing (ICASSP), 2015 IEEE International Conference on.* IEEE, 2015, pp. 5058–5062.

[36] S. J. Pan, Q. Yang *et al.*, "A survey on transfer learning," *IEEE Transactions on knowledge and data engineering*, vol. 22, no. 10, pp. 1345–1359, 2010.

[37] C. Busso and S. S. Narayanan, "Interrelation between speech and facial gestures in emotional utterances: a single subject study," *IEEE Transactions on Audio, Speech, and Language Processing*, vol. 15, no. 8, pp. 2331–2347, 2007.

[38] Y. Kim, H. Lee, and E. M. Provost, "Deep learning for robust feature generation in audiovisual emotion recognition," in *Acoustics, Speech and Signal Processing (ICASSP), 2013 IEEE International Conference on.* IEEE, 2013, pp. 3687–3691.

[39] C. Busso, M. Bulut, C.-C. Lee, A. Kazemzadeh, E. Mower, S. Kim, J. N. Chang, S. Lee, and S. S. Narayanan, "Iemocap: Interactive emotional dyadic motion capture database," *Language resources and evaluation*, vol. 42, no. 4, p. 335, 2008.

[40] C. Busso, S. Parthasarathy, A. Burmania, M. AbdelWahab, N. Sadoughi, and E. M. Provost, "Msp-improv: An acted corpus of dyadic interactions to study emotion perception," *IEEE Transactions on Affective Computing*, no. 1, pp. 67–80, 2017.

[41] S. Khorram, M. Jaiswal, J. Gideon, M. McInnis, and E. M. Provost, "The priori emotion dataset: Linking mood to emotion detected in-the-wild," *arXiv preprint arXiv:1806.10658*, 2018.

[42] Z. N. Karam, E. M. Provost, S. Singh, J. Montgomery, C. Archer, G. Harrington, and M. G. Mcinnis, "Ecologically valid long-term mood monitoring of individuals with bipolar disorder using speech," in *Acoustics, Speech and Signal Processing (ICASSP), 2014 IEEE International Conference on.* IEEE, 2014, pp. 4858–4862.

[43] S. Khorram, J. Gideon, M. G. McInnis, and E. M. Provost, "Recognition of depression in bipolar disorder: Leveraging cohort and person-specific knowledge." in *INTERSPEECH*, 2016, pp. 1215–1219.

[44] S. O. Sadjadi and J. H. Hansen, "Unsupervised speech activity detection using voicing measures and perceptual spectral flux," *IEEE Signal Processing Letters*, vol. 20, no. 3, pp. 197–200, 2013.

[45] J. A. Russell, "Core affect and the psychological construction of emotion." *Psychological review*, vol. 110, no. 1, p. 145, 2003.

[46] Z. Aldeneh, S. Khorram, D. Dimitriadis, and E. M. Provost, "Pooling acoustic and lexical features for the prediction of valence," in *Proceedings of the 19th ACM International Conference on Multimodal Interaction.* ACM, 2017, pp. 68–72.

[47] E. M. Provost, "Identifying salient sub-utterance emotion dynamics using flexible units and estimates of affective flow," in *Acoustics, Speech and Signal Processing (ICASSP), 2013 IEEE International Conference on.* IEEE, 2013, pp. 3682–3686.

[48] "SoX, Sound eXchange (v14.4.1)," http://sox.sourceforge.net/.

[49] D. Povey, A. Ghoshal, G. Boulianne, L. Burget, O. Glembek, N. Goel, M. Hannemann, P. Motlicek, Y. Qian, P. Schwarz *et al.*, "The kaldi speech recognition toolkit," in *IEEE 2011 workshop on automatic speech recognition and understanding*, no. EPFL-CONF-192584. IEEE Signal Processing Society, 2011.

[50] D. P. Kingma and J. Ba, "Adam: A method for stochastic optimization," *arXiv preprint arXiv:1412.6980*, 2014.

[51] A. Paszke, S. Gross, S. Chintala, G. Chanan, E. Yang, Z. DeVito, Z. Lin, A. Desmaison, L. Antiga, and A. Lerer, "Automatic differentiation in pytorch," 2017.

[52] S. Khorram, Z. Aldeneh, D. Dimitriadis, M. McInnis, and E. M. Provost, "Capturing long-term temporal dependencies with convolutional networks for continuous emotion recognition," *arXiv preprint arXiv:1708.07050*, 2017.

[53] H. M. Fayek, M. Lech, and L. Cavedon, "On the correlation and transferability of features between automatic speech recognition and speech emotion recognition." in *INTERSPEECH*, 2016, pp. 3618–3622.

[54] X. Glorot, A. Bordes, and Y. Bengio, "Deep sparse rectifier neural networks," in *Proceedings of the fourteenth international conference on artificial intelligence and statistics*, 2011, pp. 315–323.

[55] J. D. Hunter, "Matplotlib: A 2d graphics environment," *Computing In Science & Engineering*, vol. 9, no. 3, pp. 90–95, 2007.

[56] R Core Team, *R: A Language and Environment for Statistical Computing*, R Foundation for Statistical Computing, Vienna, Austria, 2013. [Online]. Available: http://www.R-project.org/

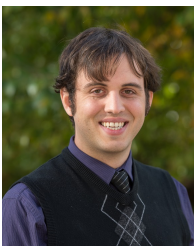

**John Gideon** is a Ph.D. candidate working with Professor Emily Mower Provost in Computer Science and Engineering at the University of Michigan, Ann Arbor. He received his B.S. in Electrical Engineering and M.S. in Computer Engineering from the University of Cincinnati, both in 2013. He is a member of IEEE, Eta-Kappa-Nu, and Tau-Beta-Pi. His research interests are the recognition of emotion and mood from speech for the improvement of medical care, as well as the design of multimodal assistive technologies for everyday tasks. He is driven by an underlying interest in human psychology and the way people perceive their interactions with one another.

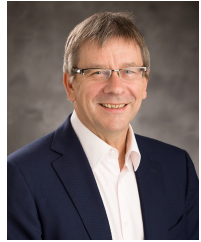

**Melvin G McInnis, MD,** is the Thomas B and Nancy Upjohn Woodworth Professor of Bipolar Disorder and Depression and Professor of Psychiatry. He is the Director of the HC Prechter Bipolar Program and Associate Director of the Depression Center at the University of Michigan. He is a Fellow of the Royal College of Psychiatry (UK) and Fellow of the American College of Neuropsychopharmacology. Dr. McInnis trained in Canada, Iceland, England, and USA, he began a faculty position in Psychiatry at Johns Hopkins University (1993) and was recruited to the University of Michigan in 2004. His research interests include the genetics of bipolar disorder and longitudinal outcome patterns in mood disorders. He has received awards recognizing excellence in bipolar research from the National Alliance for the Mentally Ill (NAMI) and National Alliance for Research in Schizophrenia and Affective Disorders (NARSAD). He has published over 250 manuscripts related to mood disorders research, and is widely engaged in collaborative research focused on identifying biological mechanisms of disease and predictive patterns of outcomes in mental health.

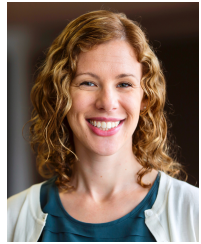

**Emily Mower Provost** is an Associate Professor in Computer Science and Engineering at the University of Michigan. She received her B.S. in Electrical Engineering (summa cum laude and with thesis honors) from Tufts University, Boston, MA in 2004 and her M.S. and Ph.D. in Electrical Engineering from the University of Southern California (USC), Los Angeles, CA in 2007 and 2010, respectively. She is a member of Tau-Beta-Pi, Eta-Kappa-Nu, and a member of IEEE and ISCA. She has been awarded a National Science Foundation CAREER Award (2017), a National Science Foundation Graduate Research Fellowship (2004-2007), the Herbert Kunzel Engineering Fellowship from USC (2007-2008, 2010-2011), the Intel Research Fellowship (2008-2010), the Achievement Rewards For College Scientists (ARCS) Award (2009-2010), and the Oscar Stern Award for Depression Research (2015). She is a co-author on the paper, "Say Cheese vs. Smile: Reducing Speech-Related Variability for Facial Emotion Recognition," winner of Best Student Paper at ACM Multimedia, 2014, and a co-author of the winner of the Classifier Sub-Challenge event at the Interspeech 2009 emotion challenge. Her research interests are in human-centered speech and video processing, multimodal interfaces design, and speech-based assistive technology. The goals of her research are motivated by the complexities of the perception and expression of human behavior.